# A New Method for Classification of Datasets for Data Mining

Singh Vijendra, Hemjyotsana Parashar and Nisha Vasudeva

Faculty of Engineering & Technology,
Mody Institute of Technology & Science, Lakshmangarh, Sikar, Rajasthan, India
d_vijendrasingh@yahoo.co.in and hemjyotsana@gmail.com

*Abstract*—— **Decision tree is an important method for both induction research and data mining, which is mainly used for model classification and prediction. ID3 algorithm is the most widely used algorithm in the decision tree so far. In this paper, the shortcoming of ID3's inclining to choose attributes with many values is discussed, and then a new decision tree algorithm which is improved version of ID3. In our proposed algorithm attributes are divided into groups and then we apply the selection measure 5 for these groups. If information gain is not good then again divide attributes values into groups. These steps are done until we get good classification/misclassification ratio. The proposed algorithms classify the data sets more accurately and efficiently.**

## I. INTRODUCTION

Humans have been manually extracting patterns from data for centuries, but the increasing volume of data in modern times has called for more automated approaches. Information leads to power and success, and thanks to sophisticated technologies such as computers, satellites, etc., we have been collecting tremendous amounts of information. Initially, with the advent of computers and means for mass digital storage, we started collecting and storing all sorts of data, counting on the power of computers to help sort through this amalgam of information. Unfortunately, these massive collections of data stored on disparate structures very rapidly became overwhelming. A variety of information collected in digital form in databases and in flat files. business transactions, scientific data, medical and personal data, surveillance video and pictures, satellite sensing, games, digital media, CAD and software engineering data, virtual worlds, text reports and memos (e-mail messages), The World Wide Web repositories[1, 4]. Early methods of identifying patterns in data include Bayes' theorem (1700s) and regression analysis (1800s). The proliferation, ubiquity and increasing power of computer technology has increased data collection and storage. As data sets have grown in size and complexity, direct hands-on data analysis has increasingly been augmented with indirect, automatic data processing. This has been aided by other discoveries in computer science, such as neural networks, clustering, genetic algorithms (1950s), decision trees (1960s) and support vector machines (1980s). Data mining is the process of applying these methods to data with the intention of uncovering hidden patterns [1,2,13]. Classification is the processing of finding a set of models (or functions) which describe and distinguish data classes or concepts. The derived model is based on the analysis of a set of training data (i.e., data objects whose class label is known). The derived model may be represented in various forms, such as classification (IF-THEN) rules, decision trees, mathematical formulae, or neural networks. A decision tree is a flow-chart-like tree structure, where each node denotes a test on an attribute value, each branch represents an outcome of the test, and tree leaves represent classes or class distributions. Decision trees can be easily converted to classification rules. Decision trees can handle high dimensional data. Their representation of acquired knowledge in tree form is intuitive and generally easy to assimilate by humans. The learning and classification steps of decision tree induction are simple and fast with good accuracy. Decision tree induction algorithms have been used for classification in many application areas, such as medicine, manufacturing and production, financial analysis, astronomy, and molecular biology. Tree-based learning methods are widely used for machine learning and data mining applications. These methods have a long tradition and are commonly known since the works of [1, 2, and 3]. They are conceptually simple yet powerful. The most common way to build decision trees is by top down partitioning, starting with the full training set and recursively finding a univariate split that maximizes some local criterion (e.g. gain ratio) until the class distributions the leaf partitions are sufficiently pure Pessimistic Error Pruning [3] uses statistically motivated heuristics to determine this utility, while Reduced Error Pruning estimates it by testing the alternatives on separate independent pruning set. In a decision tree learner named NB Tree is introduced that has Naive Bayes classifiers as leaf nodes and uses a split criterion that is based directly on the performance of Naive Bayes classifiers in all first-level child nodes (evaluated by cross-validation) an extremely expensive procedure[7]. In [6, 10] a decision tree learner is described that computes new attributes as linear, quadratic or logistic discriminate functions of attributes at each node; these are then also passed down the tree. The leaf nodes are still basically majority classifiers, although the class probability distributions on the path from the root are taken into account. A recursive Bayesian classifier is introduced in [5]. Lots of improvement is already done on decision tree induction method for 100 % accuracy and many of them achieved the goal also but main problem on these improved methods is that they required lots of time and complex extracted rules. The main idea is to split the data recursively into partitions where the conditional independence assumption holds. A decision tree is a mapping from observations about an item to conclusions about its target value [8, 9, 10, 11, and 12]. Decision trees are commonly used in operations research, specifically in decision analysis, to help identify a strategy most likely to reach a goal. Another use of decision trees is as a descriptive means for calculating conditional probabilities. A decision tree (or tree diagram) is a decision support tool that uses a tree-like graph or model of decisions and their possible consequences, including chance event outcomes, resource costs, and utility [13]. Decision tree Induction Method has been successfully used in expert systems in capturing knowledge. Decision tree induction Method is good for multiple attribute Data sets.

This paper is organized as follows. Proposed classification algorithm is briefly described in section II. Section III contains data description and result analysis. Finally, we conclude in Section IV.

## II.    PROPOSED CLASSIFICATION ALGORITHM

Inducing Classification Models is the learning of decision trees from class-labeled training tuples. Decision trees can easily be converted to classification rules. In our proposed algorithm attributes values are firstly calculated the range to attribute values then we divide these values into different no of groups (or range).

### A.    Proposed Algorithm

1. Initialize *No_of_Group*=2
2. Calculate the Information gain for each attribute
3. Select best attribute according to information gain and calculate the range of selected attribute
4. Divide the data set into *No_of_Group* according to the range
5. Call Procedure A
6. For each group
   If All_tuples in same class or classification then exit.
   Else *No_of_Group++;*
7. If all tuples are not in same class goto step 5.
8. If No_of_*Group>=MAX* then goto step 3

Then below steps are recursively performed until we get 100% or nearly 100% classification result [13]. These concept or algorithm creates 100% accurate result and tree with minimum depth or simplest decision tree for particular data set.

---

Procedure A:
  A.  Create a node N;
     (i)  if all  attributes are in the same class, C then
           mark N =leaf node with C labeled.
     (ii) if attribute list=0 then
           mark N =leaf node with D labeled.
  B.  Attribute selection method (D, attribute list)

$$Info(D) = -\sum_{i=1}^{m} p_i \log_2 (p_i)$$

$$info_A(D) = \sum_{j=1}^{v} \frac{|D_j|}{|D|} \times Info(D_j)$$

(i) if discrete-valued and multi-way splits allowed then
      attribute list $i_{th}$ splitting attribute
(ii) For each $j_{th}$ of splitting criterion be the set of data
      tuples in D satisfying outcome j;
      if $D_j$ is empty then attach a leaf node and
      labeled with the majority class in D to node N

   Else

   Attach the node returned by generate decision
   tree ($D_j$, attribute list) to node N

   End if

   return N

  End for

## III.    EXPERIMENTAL RESULTS

The implementation of both the algorithms is performed on java. All experiments were run on a PC with a 2.0GHz processor and 1GB RAM. We create interface for proposed system for classification of data set by using Java Swing.

In this paper, different data sets are used. Various data sets are tested. Performance of the proposed algorithm is very good for some data set but some data set values are so different that entropy is not that good. We tested the proposed algorithm over real data and some synthetic data. The comparison results with ID3 are shown in fig 2, fig 3, and fig 4.

Data set 1: IRIS Data set consists of 150 data points with 3 different classes.

Data set 2: Hurricane data set consists of 50 data points with 2 different classes.

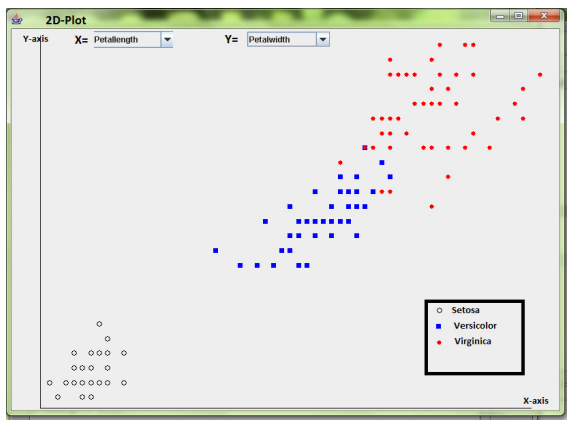

Figure 1. Classification by ID3 of Iris Data set

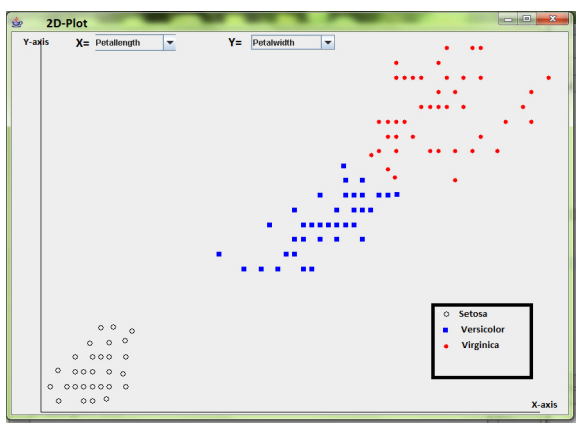

Figure 2. Classification by proposed algorithm of Iris Data set

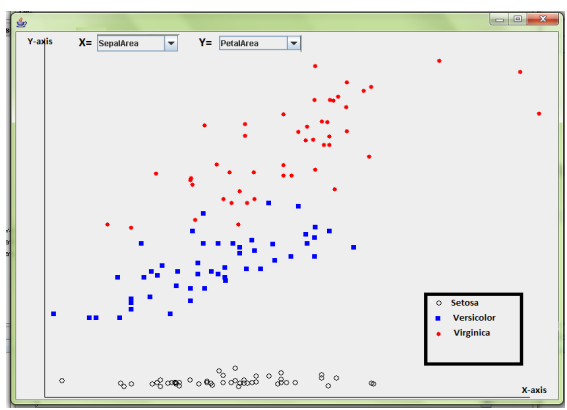

Figure 3. Classification by ID3 of Iris Data set

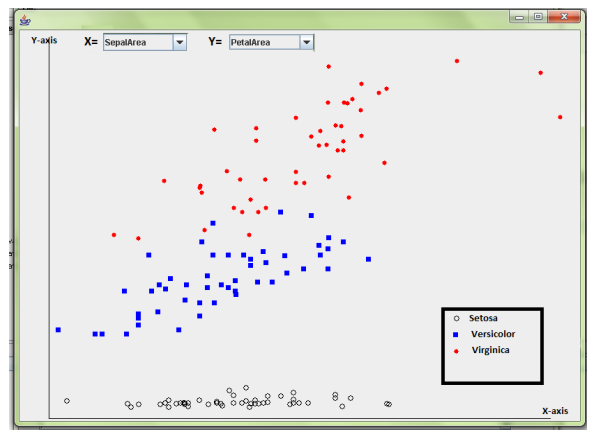

Figure 4. Classification by proposed algorithm of Iris Data set

The proposed algorithm creates a decision tree of each data sets. The Decision tree for iris data sets is shown in fig 5.

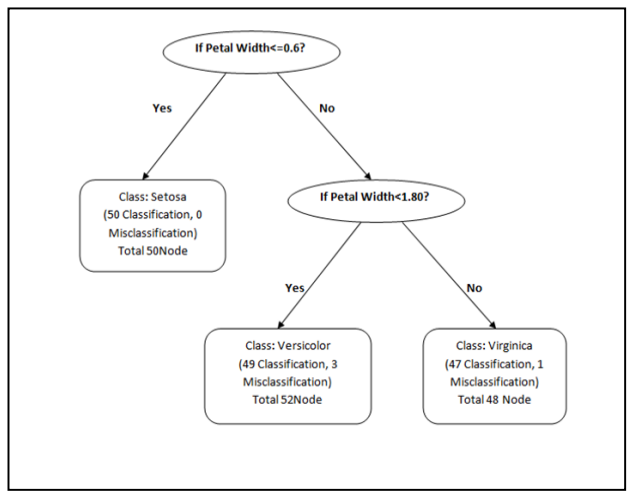

**Figure 5.** Decision Tree of Iris Data set by proposed algorithm

## IV. CONCLUSION

In this paper, we proposed a new method for classification using Decision tree. The proposed method creates decision tree and then extract rules for classification more efficiently then the pervious methods. It also improves the quality of solution and classify the data more accurately then ID3 and C4.5.